# Design and control of dynamical quantum processes
# in *ortho-para* H$_2$ conversion on surfaces
# -From atomic-scale understanding to meter-scale application-

**Rifki Muhida\*, Riza Muhida[1], Wilson A. Diño[2], Hiroshi Nakanishi[2], Hideaki Kasai[2]**

\*Department of Science in Engineering, International Islamic University Malaysia,
Jalan Gombak Kuala Lumpur 53100 Malaysia
e-mail: rifki@iiu.edu.my

[1]Department of Mechatronics Engineering, International Islamic University Malaysia,
Jalan Gombak Kuala Lumpur 53100 Malaysia
e-mail: muhida@iiu.edu.my

[2]Department of Precision Science & Technology and Applied Physics, Osaka University,
2-1 Yamadaoka, Suita, Osaka 565-0871, Japan.
e-mail: wilson@dyn.ap.eng.osaka-u.ac.jp, hiro@dy.ap.eng.osaka-u.ac.jp, kasai@dyn.ap.eng.osaka-u.ac.jp

**Abstract**

We present here a novel, cost-effective method for increasing and controlling the *ortho-para* H$_2$ (*o-p* H$_2$) conversion yield. First, we invoke two processes derived from fundamental, surface science insights, based on the effect of molecular orientation on the hydrogen-solid surface reaction, i.e., dynamical quantum filtering and steering, and apply them to enhance the *o-p* H$_2$ conversion yield. Second, we find an important factor that can significantly influence the yield i.e., inhomogeneity of spin density distribution. This factor gives us a promising possibility to increase the yield and to find the best catalyst e.g., design of materials that can function as catalysts for the *o-p* H$_2$ conversion.

## 1 Introduction

Now, issues on global warming, Arctic meltdown, and disappearance of rich ecosystems become central issue in the world, especially after Kyoto's protocol [1]. It has been explained that the temperature increases faster in the last two decades [2]. This temperature change causes the Arctic ice pack to melt at a big rate, indicating that since the 1950s, the thickness of summer polar ice shows a 40-percent reduction [2]. To resolve these global problems we must give up a carbon-based technology which is not compensated by photosynthesis and simple inefficient combustion practices which cause an emission of NO$_x$, SO$_x$ and other fragments of hydrocarbons [2]. The world currently relies on natural gases, oil, and coal as main sources of electricity generation. Within the next century, if energy consumers continue to rely on resources such as coal, petroleum, and other fossil fuels at the current rate, society will deplete its fuel reserves, specifically petroleum, then we are headed towards a global energy crisis. The energy crisis causes panic concerning not only electrical power shortages but environmental issues too.

Alternative power sources become necessary to maintain current standards of living based on power sources. The problem involves ensuring global preparedness when low energy reserves mandate the use of renewable energy resources through proper allocation of renewable energy technology, ethical support from the public and the government, financial flexibility, and environmental sustainability. This reliance requires an energy source both efficient and long-term. Efficiency includes effectiveness of production while watching out for the environment too. Hydrogen is one solution to overcome the global problems.

Hydrogen is probably the most important of all elements, and it is in abundance in the universe [3]. From an environmental point-of-view, hydrogen has been attracting a lot of attention. With water as the only emission from hydrogen combustion, hydrogen is being promoted as the energy source of the future. In order to support this developing hydrogen economy [4], infrastructures have to be built. Development of efficient processes for hydrogen extraction, and efficient processes and materials for hydrogen storage [5] would also be necessary. Thus, from an economics point-of-view, the transition to an economy based on hydrogen (energy) could, in the long run, also serve as a key to solving the problems we are currently facing.

We will give several examples to give us illustration on hydrogen technology problems and point out where is our position to overcome one of the problems. The first problem is on hydrogen production. Hydrogen can not be produced directly by drilling a well or mining. It must be extracted chemically from hydrogen-rich materials such as natural gas, water or plant matters. Large-scale production techniques that are now being used include steam filtering from natural gas, clean up of industrial by-product gases and electrolysis of water. A number of other technologies being studied include production of hydrogen from water or biomass using solar or other forms of renewable energy. Second problem is in the transportation of hydrogen gas/liquid. Hydrogen is very explosive when mixed with oxygen. In its pure state, it poses no explosive threat so is perfectly safe to ship in pipelines. Although the use of liquid hydrogen as a fuel source has a great potential, (particularly on jet airplanes and the future vehicle) the technical problems associated with the storage and delivery have not yet been overcome. In this article we will concentrate on how to overcome the storage problem resulting from *ortho-para* H$_2$ (*o-p* H$_2$) conversion. Through this study, we hope that we can reduce the cost of storing liquid hydrogen [6]. We hope that through systematic theoretical studies, we can present a simple, cost-effective method for solving a timely and urgent problem in the hydrogen technology.

The idea of using hydrogen as fuel is not a new one, but interest in it has grown in recent years, as a result of the intense efforts of the United States Government through the Department of Energy (DOE) to realize and implement hydrogen fuel technology for hydrogen-powered vehicle [7]. One important facet of this growing hydrogen technology is concerned with the storage of liquid hydrogen. When we store the hydrogen fuel, e.g., in the hydrogen fuel stations, in the rockets for space programs, in the fuel tanks of an automobile to run for 300 to 400 miles, we have to concentrate the hydrogen into a small volume. That can be done by cooling the hydrogen to an extremely low temperature or by compressing it under very high pressure as liquid. In the storage tank of liquid hydrogen, some particular procedures (e.g., extra/expensive refrigeration equipment/system) are required to keep the composition (proportion) of the two types of hydrogen molecules, known as orthohydrogen (*o*-H$_2$) and parahydrogen (*p*-H$_2$) [8]. Without the particular procedure, *o*-H$_2$ in the storage tanks would slowly but spontaneously convert to *p*-H$_2$ over a period of days or weeks, releasing enough heat to evaporate most of the liquid [7, 9]. For example, when cooling and liquefying a normal mixture (25% *p*-H$_2$, 75 % *o*-H$_2$) of hydrogen, about 40 % of the original content of the tank evaporates after 100 hours of storage [6]. In order to limit the boil-off to low levels, it is necessary to fill the tank with liquid hydrogen that has already been converted to an equilibrium composition close to 100 % *p*-H$_2$ [6, 7, 9].

The existence of the types of hydrogen is a quantum mechanical phenomenon involving the concept of spin, Pauli principle and Fermi statistics [8,10,11]. In hydrogen molecules, the three (symmetric) nuclear spin states ($|\uparrow\uparrow\rangle$, $|\uparrow\downarrow\rangle+|\downarrow\uparrow\rangle$, $|\downarrow\downarrow\rangle$) are associated with odd *j* levels, to give the *o*-H$_2$ (*j* is the rotational quantum number). The one (antisymmetric) state ($|\uparrow\downarrow\rangle-|\downarrow\uparrow\rangle$) is associated with even *j* levels, to give *p*-H$_2$. At the lowest *o*-H$_2$ and *p*-H$_2$ states the energy difference has a value of 15.08 meV [8,10]. At *T*=0 K, only the *j*=0 rotational state is occupied, so at thermal equilibrium at low temperatures a sample of hydrogen gas is pure *p*-H$_2$. At high temperatures, the gas is a mixture of both forms in the ratio of 3 parts *o*-H$_2$ to 1 part *p*-H$_2$. The two forms have different rotational contributions to the heat capacity on account of the difference in availability of rotational states. The two forms of H$_2$ also differ in their magnetic properties, in addition to their difference in thermal properties [9]. The realignment of nuclear spins is slow in the absence of dissociative adsorption or paramagnetic materials, and the 3:1 mixture persist for long periods at low temperatures.

There are generally two methods of inducing an *o-p* H$_2$ transition. One method consists of dissociating H$_2$ and then inducing the two H atoms to recombine. When dissociated, the nuclear spin states of the two atomic nuclei are no longer orientationally restricted by rotational states of the H$_2$. Upon recombination, H$_2$ is formed according to the equilibrium energy distribution determined by the temperature of the system. The other method involves the interaction between a magnetic field, produced by some catalysts e.g., magnetic material [9,11]. The magnetic field is applied along the axis of rotation of the nuclear spin, such that the external field causes a reversal of spin in one of the nuclei. This spin reversal is equivalent to an *o-p* H$_2$ transition [9]. In order to induce an *o-p* H$_2$ conversion by the catalyst, the H$_2$ must be brought close enough to the magnetic material/catalyst [9]. Without an *o-p* H$_2$ conversion catalyst, extra refrigeration equipment is required to remove the heat generated by natural *o-p* H$_2$ conversion. This extra refrigeration equipment will increase the cost of storing liquid hydrogen. The method to induce and enhance the *o*-H$_2$ to *p*-H$_2$ (*o-p* H$_2$) conversion is thus an essential in the hydrogen technology. In this article we introduce a new method to increase the *o-p* H$_2$ conversion yield on the catalysts surfaces by using molecular orientation dependence (steric effect (SE)) of *o-p* H$_2$ conversion yield (chapter 2) [12-15] and inhomogeneity of spin density distribution to finding the best catalyst for *o-p* H$_2$ conversion (chapter 3) [16]. In chapter 4 we introduce a reaction design of *o-p* H$_2$ conversion [17-20] to enhance the yield using dynamical quantum filtering (DQF) process in the associative desorption of H$_2$ from surfaces [21-23] and steric effect (SE). We summarize our results and conclusions in chapter 5.

## 2. Increasing the yield using molecular orientation (steric effect)

In this chapter, we will show how the molecular orientation of H$_2$ plays an important role in influencing the *o-p* H$_2$ conversion yield. To demonstrate this, in this article we consider the molecular orientation dependence of *o-p* H$_2$ conversion yield on Fe(OH)$_3$ catalyst. Fe(OH)$_3$ is known to be a good catalyst for the *o-p* H$_2$ conversion [11,24,25] and has been used widely in the industry to overcome problems related to storing liquid hydrogen [11,25]. Although the experiment of *o-p* H$_2$ conversion on Fe(OH)$_3$ catalyst has been done by Buyanov in 1960 [24], the detailed mechanism as to how this material can induce *o-p* H$_2$ conversion has not yet been understood. One major hurdle to a theoretical explanation of the mechanism of o-p $H_2$ conversion in Buyanov's experiments is that the structure of the $Fe(OH)_3$ could not be determined using X ray and electron photographs [11,24]. Efforts to determine the structure of $Fe(OH)_3$ are now being done by many researchers [26-28].

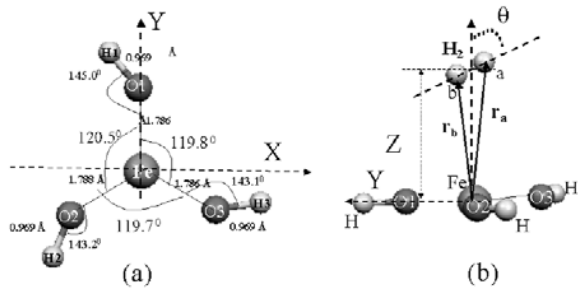

**Figure 2.1.** (a) The structure of $Fe(OH)_3$. The structure is obtained by optimization calculation using GAUSSIAN 03. (b) The relative configuration between interacting $H_2$ and $Fe(OH)_3$. The $H_2$ is sitting on the top of the Fe ion of $Fe(OH)_3$ with the $H_2$ center-of-mass (CM) distance $Z$ from the $Fe(OH)_3$ surface and $\theta$ is its molecular orientation angle with respect to the surface normal. $r_a$ and $r_b$ are the position vectors of protons a and b with respect to the Fe ion, respectively.

In the last ten to twenty years, we have seen great progress in computational physics. Through first principles calculations, we can now determine the structure of many materials and directly relate these to experimental results with a high degree of accuracy. In particular, it is now possible to theoretically explore the structure of $Fe(OH)_3$ and even simulate catalytic processes involving its interaction with $H_2$, e.g., the o-p $H_2$ conversion process.

The stable structure of $Fe(OH)_3$ [Fig. 2.1(a)] and the relative configuration between interacting $H_2$ and $Fe(OH)_3$ is shown in Fig. 2.1 (b)[51]. We assume that $H_2$ sits on top of the Fe atom in $Fe(OH)_3$ and $\theta$ is the $H_2$ molecular orientation angle with respect to $Fe(OH)_3$ surface normal, while $Z$ is the $H_2$ center-of-mass (CM) distance from the $Fe(OH)_3$ surface. The stable structure of $Fe(OH)_3$ is determined by an optimization calculation using GAUSSIAN 03 [29] via a series of density functional theory (DFT) [30] based total energy calculations, adopting Becke's three-parameter functional [31], Perdew and Wang's gradient-corrected correlation functional [32], and Dunning [33], Hay and Wadt's basis sets [34]. The obtained structure of $Fe(OH)_3$ is comparable with previous results [35]. The ground state of $Fe(OH)_3$ has $S=5/2$ ($S$ is the total spin).

We have examined the $\langle \mathbf{S} \cdot \mathbf{S} \rangle = \hbar$ values ($\mathbf{S}$ is the electron spin operator), and compared them to the $S(S+1)$ values for each total spin of the $Fe(OH)_3$ system, to check for spin contamination. In the ground state of the $Fe(OH)_3$ system the $S(S+1)$ value is 8.76 [51]. This value has the spin contamination around 0.1% where it is much smaller than the maximum value that can be allowed, i.e., 10% [36].

The molecular orientation dependence of the o-p $H_2$ conversion yield of $H_2$ on $Fe(OH)_3$ is investigated using the anti-symmetric part of the hyperfine contact interaction [11,12]. We can write the anti-symmetric part of the hyperfine contact interaction between each electron spin and two $H_2$ nuclear spins by

$$H_{HC}(Z,\theta) = \lambda_C \sum_{\alpha=1}^{N} \mathbf{S}_\alpha \cdot (\mathbf{I}_a - \mathbf{I}_b)[\delta(\mathbf{r}_a - \mathbf{r}_\alpha) - \delta(\mathbf{r}_b - \mathbf{r}_\alpha)] \quad (2.1).$$

Here, $\mathbf{I}_a$ and $\mathbf{I}_b$ are the operators for the nuclear spin of the two protons a and b of $H_2$. $\delta(\mathbf{r}_a-\mathbf{r}_\alpha)$ and $\delta(\mathbf{r}_b-\mathbf{r}_\alpha)$ represent the Dirac operators. $\mathbf{r}_n (n=a,b)$ and $\mathbf{r}_\alpha$ are the proton and electron coordinates (the position vectors) with respect to the M ion, respectively. $\mathbf{S}_\alpha$ is the electron spin operator and $\lambda_C$ is the hyperfine contact constant.

The o-p $H_2$ conversion yield as functions of the $H_2$ CM distance with respect to Fe ion and the $H_2$ molecular orientation angle with respect to the surface normal $\theta$ is proportional to the hyperfine contact contribution [12] and can be written as

$$W(Z,\theta) \propto |\xi(Z,\theta)|^2, \quad (2.2)$$

where

$$\xi(Z,\theta) = \langle \psi_{el}(\mathbf{r}_\alpha,Z,\theta)|\langle \chi_p|H_{HC}(Z,\theta)|\chi_o\rangle|\psi_{el}(\mathbf{r}_\alpha,Z,\theta)\rangle \quad (2.3)$$

$\xi(Z,\theta)$ is the hyperfine contact interaction that contributes to the o-p $H_2$ conversion yields. $\psi_{el}(\mathbf{r}_{el},Z,\theta)$ is the unperturbed wave function for the electron system obtained from GAUSSIAN 03. The $H_2$ nuclear spin wave function of the o-$H_2$ can be expressed as

$$|\chi_o\rangle = \begin{cases} |\uparrow\uparrow\rangle \\ \frac{1}{\sqrt{2}}(|\uparrow\downarrow\rangle + |\downarrow\uparrow\rangle) \\ |\downarrow\downarrow\rangle \end{cases}$$ , (2.4)

while the $H_2$ nuclear spin wave function of the p-$H_2$ can be expressed as

$$|\chi_p\rangle = \frac{1}{\sqrt{2}}(|\uparrow\downarrow\rangle - |\downarrow\uparrow\rangle)$$ . (2.5)

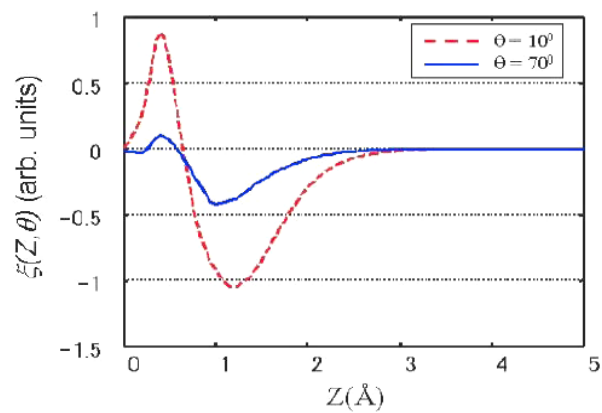

**Figure 2.2.** The matrix element of the hyperfine contact interaction $\xi(Z,\theta)$ for $\theta=10°$ and $\theta=70°$ as a function of the $H_2$ center-of-mass (CM) distance from the $Fe(OH)_3$ surface.

Fig. 2.2 shows the matrix element of the hyperfine contact interaction $\xi(Z,\theta)$ for $\theta =10°$ and $\theta =70°$. We can see that $\xi(Z,\theta)$ for $\theta =10°$ is larger than that for $\theta =70°$. The strong dependence on the $H_2$ molecular orientation angle $\theta$ is due to the antisymmetric part of the hyperfine contact interaction in Eq. (2.1) [52]. The maxima and minima for $\theta =10°$ are found at $Z=0.5$ Å and $Z=1.2$ Å, respectively, whereas the maxima and minima for $\theta =70°$ are found at $Z=0.5$ Å and $Z=1.0$ Å, respectively. These extrema correspond to the maximum contacts between the nuclear spin of $H_2$ and the electron spin density distribution of $Fe(OH)_3$-$H_2$ system[52].

The o-p $H_2$ conversion yield has shown that a perpendicular orientation is preferred to a parallel orientation. We expect that the behavior of the o-p $H_2$ conversion yield like this is a general behavior, as we have shown in the previous studies on the o-p $H_2$ conversion on metal oxide surfaces [12-14] and metal surfaces [15]. These results will give us steric effect (SE) where o-p $H_2$ conversion yield for $H_2$ with a $H_2$ doing cartwheel-like rotation (CLR $H_2$) is higher than that for a $H_2$ doing helicopter-like rotation HLR $H_2$ [14]. These results indicate that there is a promising possibility to enhance the o-p $H_2$ conversion yield by using steric effect (SE).

## 3  Increasing the yield using inhomogeneity of spin density of surface catalyst

We are now extending this study to investigate other factors that can influence the yield, e.g., inhomogeneity of spin density distribution. To demonstrate how the inhomogeneity of spin density distribution can influence the o-p $H_2$ conversion yield we consider the interaction of $H_2$ with new magnetic material clusters, i.e., organometallic compounds of multiple-decked sandwich clusters viz., $M(C_6H_6)_2$ (M=Mn, Fe, Co) that have been synthesized experimentally, e.g., by Cloke et al. [37] and Kaya's group [38]. Also, they have become the subject of many studies for the past several decades.

The electronic structures of these materials have been investigated by Miyajima et al. [39], and it has been shown that these materials exhibit semiconducting and even conducting properties in the solid states [39]. The magnetic properties of these novel materials were investigated for the first time by Yasuike et al.[40], when they investigated the spin state of $V_2(C_6H_6)_3$ system. However, they could not exactly determine which state is more stable, i.e., triplet or singlet. Pandey et al., [41] calculated the spin multiplicity of $M_n(C_6H_6)_{n+1}$ (M=Sc, Ti, V, Cr, Mn, Fe, Co, Ni, and n = 1,2) systems, and obtained agreements with electron spin resonant experiments done by Cloke et al. [42]. We have investigated the spin polarization and have determined local magnetic moments of the transition metal atom M and the benzene, in the $M(C_6H_6)_2$ clusters [43-45].

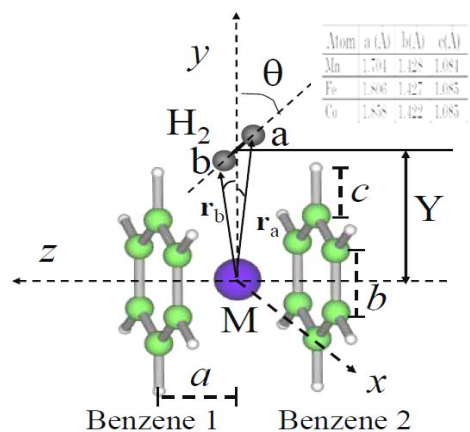

**Figure 3.1.** Model system of an interaction between $H_2$ and multiple-decked sandwich clusters of $M(C_6H_6)_2$ (M=Mn,Fe,Co). $H_2$ impinges on the atom M with its CM position Y along the y-axis ($H_2$ is in the xy-plane), as shown in the figure. $\mathbf{r}_a$ and $\mathbf{r}_b$ denote the position vectors of proton a and b with respect to the M, respectively. The $H_2$ molecular orientation angle with respect to the surface normal is given by $\theta$. Here, a is the distance from the transition metal atom to the center of benzene plane. b is the distance between the two C atoms in benzene, and c is the distance between the C atom and the H atom in benzene. The benzene ring is parallel to the xy-plane while the z-axis is perpendicular to the plane of benzene. The a, b and c values for Mn, Fe and Co systems are shown in the Table.

The model we use to investigate the o-p $H_2$ conversion on $M(C_6H_6)_2$ clusters is shown in Fig. 3.1[16]. We assume that $H_2$ impinges on the atom M with its CM position Y lying along the y-axis ($H_2$ is in the xy-plane). The molecular orientation angle with respect to the y-axis is given by $\theta$. Optimization of the $M(C_6H_6)_2$ structure is obtained by performing a series of density functional theory (DFT) that has been formulated in chapter 2 using GAUSSIAN 03.

We have checked the existence of spin contamination in this system, e.g., $Fe(C_6H_6)_2$ in the ground state, the $<\mathbf{S}.\mathbf{S}>=\hbar$ value takes 2.0641, while $S(S+1) = 2,0$. This value is much smaller than the maximum value that can be allowed, 10% [36]. To compare with previous studies by Pandey et al. [41], we have also considered the $V(C_6H_6)_2$ system. We have found that the magnetic moment of $V(C_6H_6)_2$ system is 1.0 Bohr magneton ($\mu_B$) corresponding to the spin multiplicity 2.0. This multiplicity value agrees with the experiment by Cloke et al. [42] and it also agrees with previous studies by Pandey et al. [41]. The local magnetic moment of M in $M(C_6H_6)_2$ system has been determined in the previous studies [43-45]. The magnetic moments of Mn,

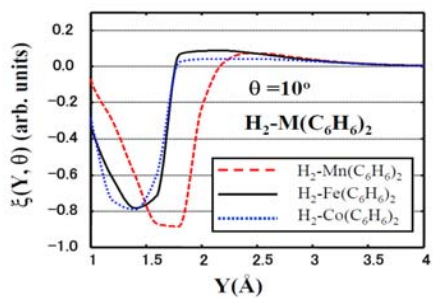

**Figure 3.2.** The hyperfine contact contribution as a function of the $H_2$ CM distance from the atom M of $M(C_6H_6)_2$ clusters (M = Mn, Fe, Co). The $H_2$ molecular orientation angle with respect to the surface normal $\theta = 10^0$.

Fe and Co in $M(C_6H_6)_2$ system are found 1.1695 $\mu_B$, 2.3816 $\mu_B$ and 1.4273 $\mu_B$, respectively [43]. These magnetic moments are lower than isolated case for Mn, Fe and Co ions. Decreased magnetic moments in M in these materials have been discussed in the references [43, 44]. When $H_2$ impinges on the $M(C_6H_6)_2$ system, e.g., when $H_2$ CM distance from the atom M is 1.8 Å and the $H_2$ molecular orientation angle $\theta = 10^0$, the magnetic moment of Mn in $Mn(C_6H_6)_2$ system becomes 3.1 $\mu_B$. For the $Fe(C_6H_6)_2$ and $Co(C_6H_6)_2$ systems the magnetic moments become 2.3 $\mu_B$ and 1.3 $\mu_B$, respectively. The increases in the magnetic moments are attributed to the $H_2$. For example, when the $H_2$ CM position is 1.8 Å from the atom M, the local magnetic moments of atoms of $H_2$ in the $Mn(C_6H_6)_2$ system increase to 0.068 $\mu_B$ and -0.0034 $\mu_B$, for atoms a and b, respectively.

To determine the o-p $H_2$ conversion yield, we use the previous method that was explained in chapter 2. In Fig. 3.2 we show the hyperfine contact contribution $\xi(Y,\theta)$ for the $H_2$ molecular orientation angle $\theta = 10^0$ as a function the $H_2$ CM distance from the atom M. The behavior of those curves is due to the spin interaction between the two protons of $H_2$ and the electron spin density of $H_2$-$M(C_6H_6)_2$ systems. It can be seen that the dip in $\xi(Y,\theta)$ for $H_2$ on $Mn(C_6H_6)_2$ system is larger than those of $H_2$ on $Fe(C_6H_6)_2$ and $Co(C_6H_6)_2$ systems. To explain this, we show the spin density distribution of the $H_2$-$Mn(C_6H_6)_2$ [Fig. 3.3(a)], the $H_2$-$Fe(C_6H_6)_2$ [Fig. 3.3(b)] and the $H_2$-$Co(C_6H_6)_2$ [Fig. 3.3(c)] systems when the $H_2$ CM distance from the atom M is 1.8 Å. Inhomogeneity of spin density distribution along the y-axis for the $H_2$-$Mn(C_6H_6)_2$ system is the largest, followed by the $H_2$-$Fe(C_6H_6)_2$ and $H_2$-$Co(C_6H_6)_2$ systems. This inhomogeneity can be known by checking spin density distribution at the $H_2$ nuclear positions of those systems. For the $H_2$-$Mn(C_6H_6)_2$ system, the spin density distribution at $r_a$ and $r_b$ are -0.0688 and 0.0225, respectively, resulting to net -0.0912 beta spin density. For the $H_2$-$Fe(C_6H_6)_2$ system, the spin density at $r_a$ and $r_b$ are -0.01466 and -0.0008, respectively, resulting to net 0.0155 alpha spin density. For the $H_2$-$Co(C_6H_6)_2$ system, the spin density at $r_a$ and $r_b$ are -0.00954 and -0.00032, respectively, resulting to net 0.0092 alpha spin density. Although the local magnetic moment of Fe is the largest in $Fe(C_6H_6)_2$ system, the inhomogeneity of spin density distribution along the y-axis of $H_2$-$Fe(C_6H_6)_2$ system material is lower than that of the $H_2$-$Mn(C_6H_6)_2$ system.

Increasing magnetic moments that followed by increasing conversion yield has been clarified experimentally by Ashmead et al. [46] in 1964, and Selwood [47,48] in 1970. Our results also show a strong dependence on the $H_2$ CM distance from M (Fig. 2). For the distances far from M, the conversion yield is small while when $H_2$ gets the closer to M (e. g, in the physisorbed $H_2$) the yield increases. This behavior has agreement with experiments of the o-p $H_2$ conversion on surface catalysts [49].

This result gives us a promising possibility to increase this conversion by finding the best catalyst e.g., design of materials that can function as catalysts for the o-p $H_2$ conversion. Inhomogeneity of spin density distribution thus can be considered as an important factor to look for the best catalyst for the o-p $H_2$ conversion.

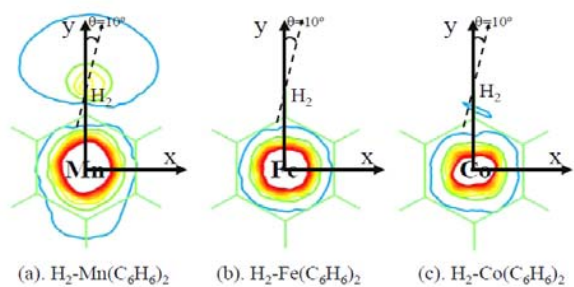

**Figure 3.3.** Two-dimensional contour (xy-plane) of spin density distribution when $H_2$ interacts with $M(C_6H_6)_2$ (M=Mn,Fe,Co). The $H_2$ CM distance from the atom M is 1.8 Å and the $H_2$ molecular orientation angle $\theta = 10^0$. (a) Spin density distribution of $H_2$-$Mn(C_6H_6)_2$ system, (b) spin density distribution of $H_2$-$Fe(C_6H_6)_2$ system, (c) spin density distribution of $H_2$-$Co(C_6H_6)_2$ system.

## 4   Reaction Design of *ortho-para* $H_2$ Conversion

In chapter 2 we have shown a promising possibility to increase the o-p $H_2$ conversion yield using molecular orientation of $H_2$ on catalyst surfaces. This molecular orientation will give us steric effect [14], where the o-p $H_2$ conversion yield for a $H_2$ doing cartwheel-like rotation (CLR $H_2$) shows a larger value than that for a $H_2$ doing helicopter-like rotation (HLR $H_2$). Based on this effect, we can increase the o-p $H_2$ conversion yield if we orient the rotational axis of $H_2$ parallel to the catalyst surface. The problem now is how to control the $H_2$ orientation so that we only have a CLR $H_2$ with respect to the surface (i.e., the $H_2$ rotational axis is parallel to the surface). This could be achieved through the dynamical quantum filtering (DQF) process via associative desorption [21-23].

Diño et al. [21-23], when they investigated the orientation-dependence of H$_2$ interaction with Cu(001) found that metal surfaces can, via the orientation-dependence of potential energy surface (PES), act or be utilized as rotational quantum state filters for molecules and induce, e.g., desorbing H$_2$ to exhibit rotational alignment. They showed that the resulting alignment of desorbed molecules, as determined by the value of quadrupole alignment factor $A_o^{(2)}(j)$, exhibits a nonmonotonic dependence on the rotational quantum number $j$ and the translational energy $E_t$. From the calculation of desorption probability [21-23], they found the corresponding quadrupole alignment factors

$$A_o^{(2)}(E_{tot}) = \frac{\sum_{m_j}[3m_j^2 - j(j+1)]D_{mnjm_j}^{m'n'}(E_{tot})}{\sum_{m_j}j(j+1)D_{mnjm_j}^{m'n'}(E_{tot})} \quad (4.1)$$

Here, $m$, $n$, $m'$ and $n'$ are the quantum numbers for the surface parallel translational motion of H$_2$. $j$ and $m_j$ are quantum numbers for the rotational motion of H$_2$ and $j$ is the absolute value of the angular momentum vector **j**, and $m_j$ corresponds to the surface normal component of the angular momentum vector **j**. The total energy of the system $E_{tot}$ is defined as the sum of the kinetic energy of the surface normal translational motion, surface parallel translational motion, rotational motion, and vibrational motion of H$_2$. $D^{m'n'}{}_{mnjm_j}(E_{tot})$ is desorption probability as functions of final states of the rotational motion ($j,m_j$), and initial ($m',n'$) and final ($m,n$) states of the surface parallel translational motion of H$_2$ on the surface.

The quadrupole alignment factor $A_o^{(2)}(E_{tot})$ can be experimentally determined by using resonance-enhanced multiphoton ionization (REMPI) and laser-induced fluorescence (LIF), and gives us information regarding the degree of alignment and orientational preference of H$_2$. It assumes values in the range [-1,3j/(j+1)-1]. A CLR H$_2$ ($m_j \approx 0$) has $A_o^{(2)}(E_{tot}) < 0$, while HLR H$_2$ ($m_j \approx j$) has $A_o^{(2)}(E_{tot}) > 0$. A spatially isotropic distribution of the angular momentum vector **j** is described by $A_o^{(2)}(E_{tot}) = 0$. In Fig. 4.1, one can see that slowly desorbing H$_2$ [i.e., the final translational energy ($E_t$) « $V_{min}$] exhibit a CLR H$_2$, while fast desorbing H$_2$ [$E_t$ » $V$min] exhibit a HLR H$_2$. $V_{min}$ is the minimum activation barrier for H$_2$ dissociation, with the H-H bond oriented parallel to the surface [$V_{min} \approx 0.5$ eV] [21-23].

Now, we propose a new method to enhance the $o$-$p$ H$_2$ conversion yield/rate of H$_2$ inter-acting with a solid surface. It consists of two steps. The first step involves the DQF process [21-23], and the second step relies on the steric effect (SE) of the $o$-$p$ H$_2$ conversion [12-15]. The purpose of the DQF process is to align the rotation of $o$-H$_2$ such that we have only a HLR H$_2$ or a CLR H$_2$. This DQF process could be performed, e.g., by permeating H atoms through some metal crystal. After permeation, the desorbing H$_2$ will be exhibiting either a CLR H$_2$ or a HLR H$_2$ depending on their final translational energy upon desorption. As shown in Fig. 4.1, if we devise a means to either select only slowly desorbing molecules or fast desorbing molecules, then we can obtain exclusively CLR or HLR H$_2$. In the second step, as shown in Fig. 4.2(a), we bring these aligned $o$-H$_2$ close to a catalyst surface. Based on the steric effect, CLR $o$-H$_2$ will be converted to $p$-H$_2$ with higher rate than HLR $o$-H$_2$.

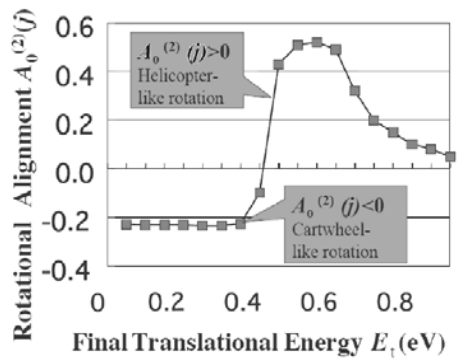

**Figure 4.1.** Rotational alignment of H$_2$ as a function of the final translational energy. Molecules with $A_o^{(2)}(j) > 0$ exhibit a HLR H$_2$, while molecules with $A_o^{(2)}(j) < 0$ exhibit a CLR H$_2$. Experiment has been done by Hou et al. [50].

To realize this idea we develop a conceptual design of a machine that will increase the $o$-$p$ H$_2$ conversion yield [17-20,53]. The $o$-H$_2$ with free rotation, first undergoes the DQF process. In the second stage, for the H$_2$ desorption with CLR, we put catalyst A in the position as shown in Fig. 4.2(b), while for the H$_2$ desorption with HLR we use catalyst B. This machine will give us higher catalytic surface area for the $o$-$p$ H$_2$ conversion and improves the efficiency of the second process. We can use this machine during the liquefaction process [17-20].

## 5   Summary and conclusions

The $o$-$p$ H$_2$ conversion process is extremely important in the industry, e.g., when considering the production, storage and utilization of hydrogen. At low temperature $o$-H$_2$ is unstable, and changes toward the more stable $p$-H$_2$ over time, liberates heat which can cause hydrogen evaporation within the storage vessel. Since $o$-H$_2$ makes up 75 % of normal H$_2$ at room temperature, they can considerably complicate the job of storing liquid hydrogen. Without an $o$-$p$ H$_2$ conversion catalyst, extra refrigeration equipment is required to remove the heat generated by natural $o$-$p$ H$_2$ conversion. The methods to induce and enhance the $o$-H$_2$ to $p$-H$_2$ ($o$-$p$ H$_2$) conversion are thus an essential in the hydrogen technology.

In this article we introduced new methods to increase the $o$-$p$ H$_2$ conversion yield on the catalysts surfaces by using:

1. *Molecular orientation.*
   We expected that the orientation of H$_2$ on surface catalyst would also play an important role in enhancing the $o$-$p$ H$_2$ conversion. As an illustration, we consider the molecular orientation dependence of *o-p* $H_2$ conversion yield on the Fe(OH)$_3$ catalyst that is known to be good catalyst for *o-p* $H_2$ conversion. By using steric effect (SE) and dynamical quantum filtering (DQF) effect in the associative desorption we propose new reaction design of the *o-p* $H_2$ conversion on catalyst surfaces to increase the yield. This design will give us higher catalytic surface area for the *o-p* $H_2$ conversion and improves the efficiency.

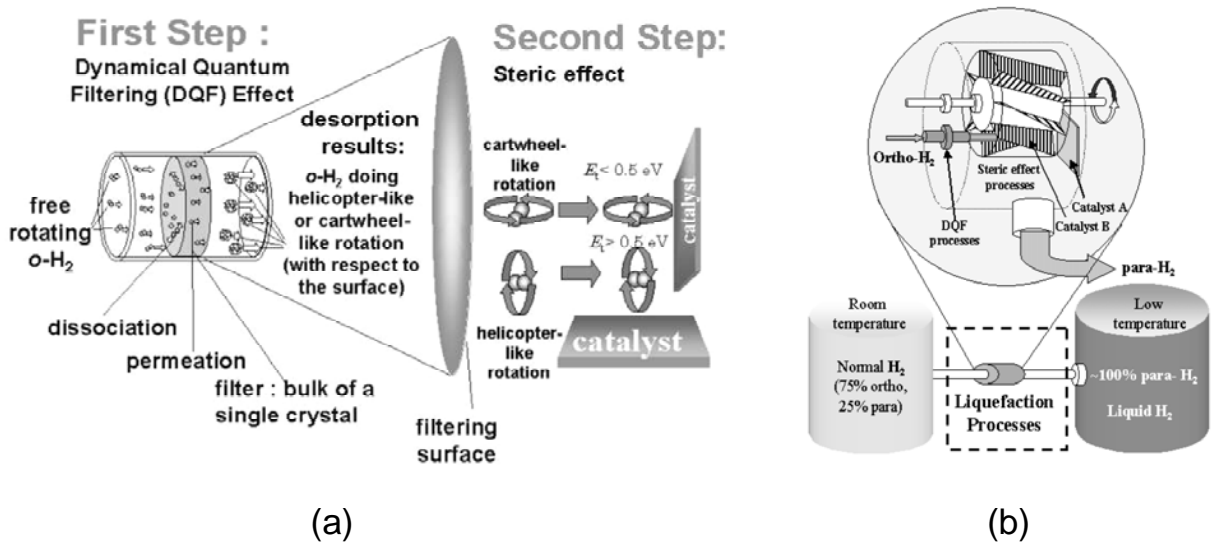

(a)                      (b)

**Figure 4.2.** (a) The new method to enhance the *o-p* $H_2$ conversion yield/rate we propose here consists of two steps. The first step involves the dynamic quantum filtering process, and the second step relies on the steric effect (SE) of the *o-p* $H_2$ conversion. The purpose of the dynamic quantum filtering (DQF) process is to align the free rotating *o*-$H_2$ such that we have only a HLR $H_2$ or a CLR $H_2$. (b) A conceptual design of a machine to realize the reaction design of the *o-p* $H_2$ conversion by using DQF process and the SE.

2. *Inhomogeneity of spin density distribution.*

Based on our series of studies on magnetic properties of nanoclusters and its interactions with $H_2$ [16,43-45], we found that the inhomogeneity of spin density distribution can influence the *o-p* $H_2$ conversion yield. This result gives us a promising possibility to increase this conversion by finding the best catalyst e.g., design of materials that can function as catalysts for the *o-p* $H_2$ conversion. Inhomogeneity of spin density distribution thus can be considered as an important factor to look for the best catalyst for the *o-p* $H_2$ conversion.

## 6 Acknowledgement

This work is partly supported by International Islamic University Malaysia and the Ministry of Education, Culture, Sports, Science and Technology of Japan (MEXT) through Grants-in-Aid for Scientific Research on Priority Areas (Developing Next Generation Quantum Simulators and Quantum-Based Design Techniques), and through Grants-in-Aid for the 21st Century Center of Excellence (COE) Program "Core Research and Advance Education Center for Material Science and Nano-Engineering" supported by the Japan Society for the Promotion of Science (JSPS) and for the New Energy and Industrial Technology Development Organization's (NEDO) Materials and Nanotechnology Program. Calculations were performed using the supercomputer facilities of the Institute for Solid State Physics (ISSP) Supercomputer Center (University of Tokyo) and the Japan Atomic Energy Agency (ITBL, JAEA).